\pdfoutput=1

\documentclass[11pt]{article}

\usepackage{EMNLP2022}
\usepackage{times}
\usepackage{latexsym}
\usepackage{amsmath}
\usepackage{graphicx}

\usepackage[T1]{fontenc}

\usepackage[utf8]{inputenc}

\usepackage{microtype}
\usepackage{booktabs}

\usepackage{inconsolata}
\usepackage{mathptmx}
\usepackage{multirow}
\usepackage{color, colortbl}
\usepackage{url}

\newcommand{\modelname}{Bipolar-encoder}
\newcommand{\dsname}{counterargs-18}

\title{Revisiting the Role of Similarity and Dissimilarity in \\Best Counter Argument Retrieval}

\author{Hongguang Shi \\
  AI school,Nanjing University \\
  \texttt{shg123789@gmail.com} \\\And
  Shuirong Cao \\
  AI school,Nanjing University \\
  \texttt{srcao@smail.nju.edu.cn} \\\And
  Cam-Tu Nguyen \\
  AI school,Nanjing University \\
  \texttt{ncamtu@nju.edu.cn} \\}

\begin{document}
\maketitle
\begin{abstract}
This paper studies the task of best counter-argument retrieval given an input argument. Following the definition that the best counter-argument addresses the same aspects as the input argument while having the opposite stance, we aim to develop an efficient and effective model for scoring counter-arguments based on similarity and dissimilarity metrics. We first conduct an experimental study on the effectiveness of available scoring methods, including traditional Learning-To-Rank (LTR) and recent neural scoring models. We then propose Bipolar-encoder, a novel BERT-based model to learn an optimal representation for simultaneous similarity and dissimilarity. Experimental results show that our proposed method can achieve the accuracy@1 of 48.86\%, which significantly outperforms other baselines by a large margin. When combined with an appropriate caching technique, Bipolar-encoder is comparably efficient at prediction time. 
\end{abstract}

\section{Introduction}



Arguments are essential for developing critical thinking, making rational choices, or justifying one’s past choices. Since the time of Socrates, argumentative discussions have been recommended for humans to acquire true knowledge by examining opposing views and removing all kinds of preconceptions \cite{damer2012attacking}. With the support of modern technologies, argument mining has been found useful in many applications, ranging from education \cite{wambsganss-etal-2021-supporting}, scientific discussion \cite{cheng-etal-2021-argument}, legal document analysis \cite{cabrio-serena-2018five-years}, and fighting over fake news \cite{orbach2020echo, lazer2018science}. Argumentation is also viewed as an essential tool towards explainable and beneficial AI \cite{vassiliades2021argumentation}.
\begin{figure}
    \centering
    \includegraphics[width=0.49\textwidth]{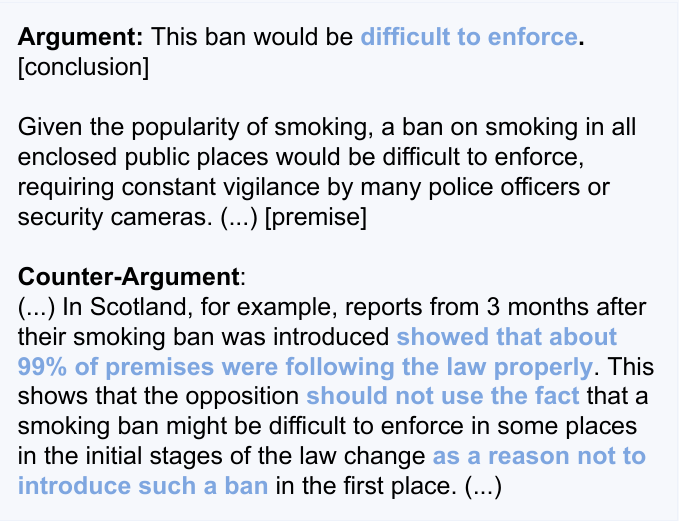}
    \caption{A sample pair of arguments from the debate with topic of ``whether should we ban smoking in public''. The example is selected from {\dsname} dataset \cite{wachsmuth2018retrieval}.}
    \label{fig:example}
\end{figure}


Central to the argumentative process is the retrieval of the best counter-argument for a given input argument. This is a very challenging task for both humans and computers, partially due to the difficulty of defining what constitutes a good (counter) argument. To date, there have been extensive studies on argument qualities both from argumentation theory \cite{walton2009argumentation-theory, damer2012attacking} and computer science \cite{tan2016winning, zhao-etal-2021-leveraging, vecchi-etal-2021-towards, emelin-etal-2021-moral}. Although we believe that qualities such as persuasion, logical, good contribution, local acceptability should play important roles in assessing counter-arguments, these qualities are difficult to formalize concretely for machines to learn. One may argue that we can avoid the formalization task by taking the data-driven approach, but it then requires a considerable labeling effort instead. 

One definition by \cite{wachsmuth2018retrieval}, which we find simple and appealing, is that the best counter-argument \textit{``invoke(s) the same aspects as the (input) argument while having the opposite stance''}. As an illustration, Figure \ref{fig:example} shows an argumentative pair where the argument is against ``banning smokes in public'' while the counter-argument supports it; both arguments discuss the same aspect of ``banning smokes is difficult to enforce''. Grounded on argumentation theory \cite{walton2009argumentation-theory}, this definition is appealing because it allows us to focus on similarity and dissimilarity to find counter-arguments without having to explicitly model topics and the topic-dependent stances. This is particularly attractive since the number of topics can be infinite. The definition is also fundamental and simple because we now can use many distance metrics, which have been studied intensively in and Machine Learning (ML), to measure ``the same aspects'' and ``the opposite stance''. Additionally, an effective method based on this definition can be later combined with other argument qualities for better counter-argument retrieval.

Taking the above definition as the starting point, \citet{wachsmuth2018retrieval} developed a simple method (Simple-SD) that heuristically combines similarity and dissimilarity metrics for scoring candidate arguments. Interestingly, on a related but slightly different task, \citet{orbach2020echo} found Simple-SD to be comparable to a model based on a now standard BERT model \cite{devlin-etal-2019-bert}. Despite such potential, the performance of simple-SD is still far from desirable. These observations lead us to several research questions: 1) Can traditional LTR methods that learn to combine various similarity and dissimilarity metrics work better than the heuristic combination in Simple-SD? 2) Can the latest neural scoring models, built on large pre-trained language models such as BERT \cite{devlin-etal-2019-bert}, perform well on this task? 3) Can we design a better model that optimizes similarity and dissimilarity for best counter-argument retrieval?

By targeting such questions in a systematic way, we make the following contributions: 
\begin{enumerate}
    \item We shed light on the effectiveness of modeling similarity and dissimilarity from different approaches.
    \item We propose {\modelname}, a novel BERT-based model that learns an optimal representation for simultaneous similarity and dissimilarity. Our model is shown to significantly outperform simple-SD and other baselines on {\dsname}, while being more effective than most of the prior approaches.
    \item We conduct an ablation study with different variants of {\modelname} and show that the integration of ``dissimilarity'' is essential for best counter-argument retrieval. 
\end{enumerate}



\section{Related Work}
Towards automatically constructing an argument to counter a given input argument, one can follow the retrieval approach or the generative approach.

\subsection{Retrieval Approach}

 Given an input argument, the retrieval approach finds arguments in a collection that best counter the input one. \citet{wachsmuth2018retrieval} defines the best counter-argument as the one that takes on the aspects of the given argument while having a different stance. The authors then proposed a simple scoring method (SimpleSD), which combines similarity metrics such as Manhattan and Earth Mover's distance. Despite its simplicity, it showcases the potential of modeling simultaneous similarity and dissimilarity for counter-argument retrieval. \citet{orbach2020echo} studied a related but slightly different task where they tried to \textit{``identify the response to a supporting speech from a set of opposing speeches, all discussion the same motion''} for fighting fake news and disinformation on social media. The main difference between the two settings is that the former works with arguments without prior information on topics or stances, while the later focuses on speeches each containing several arguments with known motion and stances. The authors reported several baselines including BERT-based classification and Simple-SD. Surprisingly, even with fine-tuning, BERT-based method did not show the advantages over Simple-SD. This probably is because an input speech is usually long and needs to be truncated for BERT, of which the input is limited to only 512 characters. As a result, we would like to focus on the setting in \cite{wachsmuth2018retrieval}, and attempt to improve the scoring model even without the topic or stance information.


In the recent years, there has been a growing interest in Argument Mining, giving rise to many techniques that can be applied for argument search. Specifically, argument extraction methods \cite{feng-hirst-2011-classifying,lawrence2019argument} can be used to recognize argumentative units (debating topics, arguments, premises) from a text collection. The relations (e.g. attack, support) between those argumentative units can be recognized with methods such as \cite{bao-etal-2021-neural, lawrence2019argument, barrow-etal-2021-syntopical, yuan-2021-interactive, cocarascu-toni-2017-identifying}. Stance detection can be utilized to detect if an argument is for or against a topic \cite{sun-etal-2018-stance, ALDayel2021stance}. Once topics, stances and argumentative relations are detected, such information can be used to facilitate best counter-argument retrieval. Here, we would like to avoid these still-challenging tasks, and assume that the information about topics and stances is not available. This is a more realistic setting because the number of topics can be infinite.

Note that in retrieval scenarios, argument qualities \cite{tan2016winning, zhao-etal-2021-leveraging, vecchi-etal-2021-towards} can be considered  for better ranking arguments and counter-arguments \cite{lorik2020quality-aware, potthast2019argument}.  Unfortunately, modeling argument qualities is difficult, and so we leave that for future considerations.

\subsection{Generative Approach}

Methods following the generative approach rely on text generation models to construct a counter argument given an input argument. \citet{le2018dave} proposed a counter argument generation based on LSTM. \citet{hua-wang-2018neural} presented a method integrating  evidences from Wikipedia to an encoder-decoder framework for generation. \citet{hua-etal-2019argument-generation} followed a two-step generation framework, where a text planning decides the talking points based on retrieved evidences, and a content realization decoder construct an informative paragraph-level argument. \citet{alshomary-etal-2021-counter} studied the problem of generating counter-argument by predicting weak premises and attacking them. \citet{al-khatib-etal-2021-employing,jo-etal-2021-knowledge-enhanced}, on the other hand, employed knowledge graphs for neural argument generation. 

Many of these counter-argument generation methods make use of a retrieval model to obtain evidences or counter-claims from a text collection. The retrieval part is also an integral component in the architecture of the Project Debater \cite{slonim2021autonomous}, the world-first debating machine. Without the retrieval component, the generated argument will be groundless, and less convincing. As a result, we believe that our method will benefit future studies in counter-argument generation as well.


\section{Background}\label{sec:background}
The problem of best counter argument retrieval can be described as follows. Assume that we have a collection of arguments $C=\{c_1,c_2,...,c_N\}$, each can be viewed as a sequence of tokens $c_i$. Given an argument point $p$, the task is to find an argument from the collection $C$ that best counters the input point based on a scoring function $f: (p,c_i)\rightarrow R$. Note that the corpus size can easily reach thousands to millions of arguments, for example, in finding counter-arguments from scientific journals or law documents. As a result, it is desirable to have an efficient filtering component $F:(p,C)\rightarrow C_{F}$ to select a candidate set $C_{F}\subset C$ where $|C_{F}|\ll N$, before applying the scoring function. We refer to this as the retrieve-and-rerank approach.

\citet{wachsmuth2018retrieval} constructed the {\dsname} dataset for best counter-argument retrieval, and defined different settings based on whether we have information about the \textit{debate theme} (e.g. culture, economy, education, etc.), the \textit{debate topic} (e.g. whether we should ban gun?), or the \textit{stance} of each argument (e.g. pro or con) on the associated topic. More specifically, in the \textbf{sdoc} setting (\textit{same debate, opposing counter}), it is assumed that topics and stances can be reliably detected and the filtering component $F$ selects only the arguments with the same topic and the different stance. Similarly, \textbf{sda} and \textbf{epc} indicate that we consider \textit{same debate arguments} or \textit{entire portal counters} as candidates. The most difficult setting is \textbf{epa} (\textit{entire portal arguments}) which assumes that we do not have any information about the debate topic or the stance. This paper focuses mostly on the \textbf{epa} setting, but also provides some results on \textbf{sdoc}, \textbf{sda} and \textbf{epc} for more insights into the effectiveness of the traditional methods having additional information such as topics or stances.

\subsection{Simultaneous Similarity and Dissimilarity}
\label{sec:simplesd}


\citet{wachsmuth2018retrieval} proposed a simple method (\textbf{Simple-SD}) that combine different similarity and dissimilarity metrics for scoring a candidate $c$ given a point $p$. Towards this end, a feature vector $\textbf{x}=\Phi(p,c)\in R^d$ is created from each point-candidate pair $(p,c)$ where $\Phi$ denotes the feature function. Correspondingly, each feature $x_i=\phi^{(i)}(p,c), i\in[1,d]$ is defined as a metric of the pair.

\paragraph{Feature Extraction} Inspired from argumentation theory \cite{walton2009argumentation-theory}, which states that a counter-argument may \textit{``target an argument's premises or conclusion''}, \citet{wachsmuth2018retrieval} decompose an argument point $p$ into two texts $\{p_{cl}, p_{pr}\}$ corresponding to the conclusion and the premise part of the argument (see Figure \ref{fig:example}). The candidate argument, however, is not decomposed in this way. As such, each feature is reformulated as  $\phi^{(i)}: agg[met(p_{cl},c), met(p_{pr},c)]\rightarrow R$, where $met$ is some metric on a pair of texts and $agg$ is an aggregation function. Simple-SD applies 4 standard aggregation functions, including \textbf{\textit{max}}, \textbf{\textit{min}}, \textbf{\textit{production}} and \textbf{\textit{summation}}. 

Depending on different choices of text representation methods and similarity/dissimilarity scores, we have many ways to define the metric function $met$. SimpleSD exploits two of such metrics: (1) \textit{Word-based Manhanttan Similarity}: the inverse Manhattan distance \cite{cha2007comprehensive} between the normalized term frequency vectors; and (2) \textit{Embedding Earth-Mover Similarity}: each text is represented by a set of word embeddings by looking up tokens in the pretrained \textit{ConceptNet Numberbatch} model \cite{speer2017conceptnet}; then the inverse Word Mover's distance \cite{kusner2015word} is calculated between the two sets of embeddings.



\begin{table}[]
    \begin{tabular}{lll}
        \toprule  Features & \multicolumn{2}{c}{Metrics} \\\cline{2-3}
          (Agg: m/x/p/s) & Text Rep. & Similarity \\
        \midrule
        $x_1, ..., x_4$ & Word-vector & Manhattan \\
        $x_2, ..., x_8$ & Embeddings & Earth Mover \\ 
        $x_{9}, ..., x_{12}$ & Word-vector & Cosine \\
        $x_{13}, ..., x_{16}$ & Bag-of-words & BM25 \\
        $x_{17},...,x_{20}$ & BERT & Cosine \\
        \toprule
    \end{tabular}
    \caption{Features extracted for a pair of arguments. Each feature is calculated by using: (1) a specific aggregation function (min(m), max(x), product(p), and sum(s)); (2) a specific text representation; and (3) a  similarity score.}
    \label{tab:features}
\end{table}

\paragraph{Scoring Function} Using 4 aggregation functions and 2 above metrics, a pair of $(p,c)$ is thus represented as a vector of 8 dimensions, corresponding to $x_1,...,x_8$ in the first two rows of Table \ref{tab:features}. A simple linear combination $f(\textbf{x})=\sum_{i=1}^8\alpha_ix_i$ is chosen as the scoring function, where a positive value of $\alpha_i$ indicates a kind of similarity, a negative one measures a kind of dissimilarity and zero means that the feature is not important and unused. \citet{wachsmuth2018retrieval} then heuristically and manually selected the values of $\alpha_i$ based on a validation set. The best result of Simple-SD on {\dsname} was reported with  $\alpha_4=\alpha_8=0.9$, $\alpha_1=\alpha_2=-0.1$ and other $\alpha_i$ being zeros. 

\subsection{Traditional LTR Models}
Simple-SD heuristically and manually probed for the combination of similarity/dissimilarity features. However, such process can  become daunting, particularly when more features are included. Accordingly, we would like to automatically infer such coefficients using machine learning. The problem is formalized as a learning-to-rank (LTR) problem, which learns the scoring function $f$ using a training dataset $\mathcal{D}=\{\langle p_i,c^+_i,c^-_{i1},c^-_{i2},...,c^-_{in}\rangle\}_{i=1}^m$ of $m$ instances. Each instance has one point $p_i$ and one correct counter-argument $c^+_i$ along with $n$ irrelevant (negative) arguments ($c^-_{ij})$. 

\paragraph{Problem Transformation} The LTR problem can be tackled in three-ways: the pointwise, pairwise and listwise approaches \cite{li2011short}. The \textit{pointwise approach} transforms the training dataset $\mathcal{D}$ to obtain $\mathcal{D}^{point}=\{(\langle p_j,c_j\rangle,y_j)\}_{j=1}^M$ where $y_j$ is $1$ if $c_j$ is the correct counter-argument for the point $p_j$, and $0$ otherwise.  The \textit{pairwise approach} constructs $\mathcal{D}^{pair}=\{\langle p_j,c^+_j, c^-_j\rangle\}_{j=1}^M$ by sampling triples of a point $p_j$, the positive counter $c^+_j$ and a (sample) negative counter $c^-_j$. The objective is to find a scoring function so that the score for the positive counter should be higher than the negative one given the input argument point. On the other hand, the \textit{listwise approach} addresses the problem directly by taking each list provided in $\mathcal{D}$ as one learning instance.  Since the pointwise approach is the most comparable to Simple-SD, we follow this approach with traditional LTR models. 

\paragraph{Feature Extraction} Following the pointwise approach, we apply feature extraction $\textbf{x}_j=\Phi(p_j,c_j)$ for each pair of $(p_j,c_j)$. The objective is to turn the problem into a traditional classification problem. Similar to SimpleSD, we also decompose a point into two parts corresponding to the conclusion and premises, and extract features of the form $\phi^{(i)}: agg[met(p_{cl},c), met(p_{pr},c)]\rightarrow R$. We use the same kinds of aggregation functions and the two metrics as in SimpleSD. In addition, we explore 3 more metrics $met$ for measuring the similarity between two texts: (1) \textit{Word-based Cosine Similarity}: the Cosine similarity between the normalized term frequency vectors of the texts; (2) \textit{Bag-of-word BM25 Score}: BM25 score \cite{robertson2009probabilistic} calculated given between texts; and (3) \textit{A BERT-based Cosine Similarity}: the Cosine similarity between the vectors of [CLS] embeddings, which are obtained by using a pretrained BERT (base, uncased) model \cite{devlin-etal-2019-bert}. All in all, we obtain 20 features as shown in Table \ref{tab:features}. 

\begin{figure}
  \centering
  \includegraphics[width=0.45\textwidth]{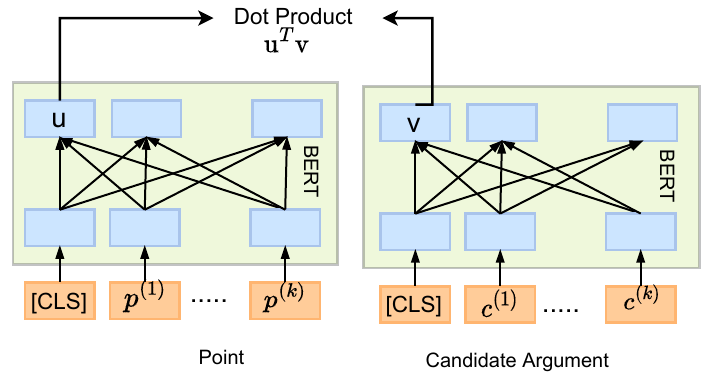}
 \caption{Bi-encoder Architecture}
 \label{fig:biencoder}
\end{figure}

\paragraph{Learning Models}  After feature extraction, the problem is transformed into learning a binary classifier on the training set $\mathcal{D}^{point}=\{(\textbf{x}_j,y_j)\}_{j=1}^M,y_j\in\{0,1\}$. In principle, many models can be used, this paper considers Logistic Regression (LR) for its simplicity and  explainability, and XGBoost \cite{chen2016xgboost} for its proven performance.

\section{Neural Scoring Models}\label{sec:neural-models}
Large pre-trained language models such as BERT have led to remarkable improvements over multiple applications \cite{devlin-etal-2019-bert}. For tasks that need pairwise comparison between texts, state-of-the-art results can be obtained by following the two commonly used architectures, Bi-encoder and Cross-encoder \cite{karpukhin-etal-2020-dense,qu-etal-2021-rocketqa,humeau2019poly}. 

\subsection{Bi-Encoder Model} 
In a Bi-encoder  (see Figure \ref{fig:biencoder}), two independent networks are used to convert a pair of texts to two dense vectors. Although  many types of networks can be applied, this work exploits pretrained BERT (base, uncased) models \cite{devlin-etal-2019-bert} as the base encoders. These encoders, denoted as $BERT_p$ and $BERT_c$, take the outputs at the characters [CLS] as the representations for the input sequences $p$ and $c$. The scoring function is then defined as the dot product of the outputs:
\begin{equation*}
dis(p,c)=\text{dot}[BERT_p(p), BERT_c(c)]
\end{equation*}

\paragraph{Training} The objective is to update the (pretrained) BERT encoders so that they produce more suitable representations for the best counter-argument retrieval. Following the pairwise approach, Bi-encoder is trained by minimizing the Triplet loss on the training dataset $\mathcal{D}^{pair}=\{\langle p_j,c^+_j, c^-_j\rangle\}_{j=1}^M$, where a loss associated with each sample $\langle p_j,c^+_j, c^-_j\rangle$ is defined as: 
\begin{equation*}
    \mathcal{L}^{(i)}_{TL}=\max(dis(p_j,c^+_j)-dis(p_j,c^-_j)+\alpha,0)
\end{equation*} 
Intuitively, we want to keep the distance $dis(p_j,c^+_j)$ smaller than $dis(p_j,c^-_j)$. Here, the distance $dis(p,c)$ is measured using  Euclidean distance of the two vectors, obtained by encoding the two input texts $p$ and $c$ with $BERT_p$ and $BERT_c$. Note that one can follow the listwise approach and train Bi-encoder with the negative log likelihood loss \cite{karpukhin-etal-2020-dense}, or the pointwise approach and train the encoders with the cross-entropy loss \cite{humeau2019poly}. 

\begin{figure}
  \centering
  \vspace{-.2cm}
  \includegraphics[width=0.42\textwidth]{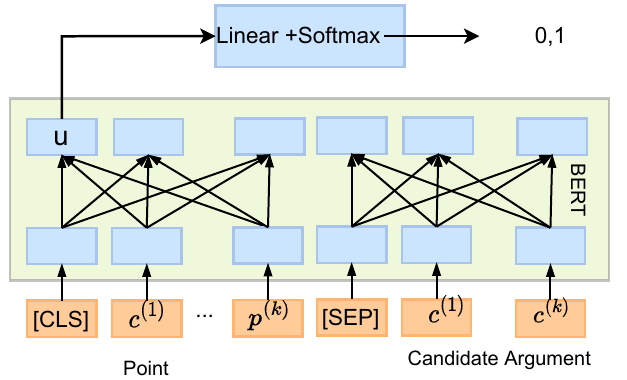}
 \caption{Cross-encoder architecture}
 \vspace{0.0cm}
 \label{fig:cross-encoder}
\end{figure} 
\subsection{Cross-Encoder Model} 
In a Cross-encoder (see Figure \ref{fig:cross-encoder}), the point and candidate are joined by the character [SEP] to form one single sequence, which is converted into a dense vector using a single encoder. Once again, we use the pretrained BERT (base, uncased) as the base encoder, and take the output representation of the character [CLS] as the embedding for the concatenated sequence. To score one candidate, a linear layer is applied to the embedding to reduce it to a vector of two dimensions, then a softmax is used to obtain $\hat{y}=f(p,c)$, the probability of $c$ being the correct counter-argument for $p$, as the final score:
$$f(p,c)=\text{softmax}(\text{linear}(BERT([p,c])))$$
where $[p,c]$ denotes the concatenation of $p$ and $c$. 

\paragraph{Training} Following the pointwise approach, we train the Cross-encoder by minimizing the Cross-entropy loss on the training dataset $\mathcal{D}^{point}=\{(\langle p_j,c_j\rangle,y_j)\}_{j=1}^M, y_i\in{0,1}$, where a loss associated with one sample is calculated by:
\begin{equation*}
\mathcal{L}^{(j)}_{CE}= -[y_j\log(\hat{y}_j)+(1-y_j)\log(1-\hat{y}_j)]
\end{equation*}

\subsection{Remarks}

\begin{figure*}
  \centering
  \includegraphics[width=0.85\textwidth]{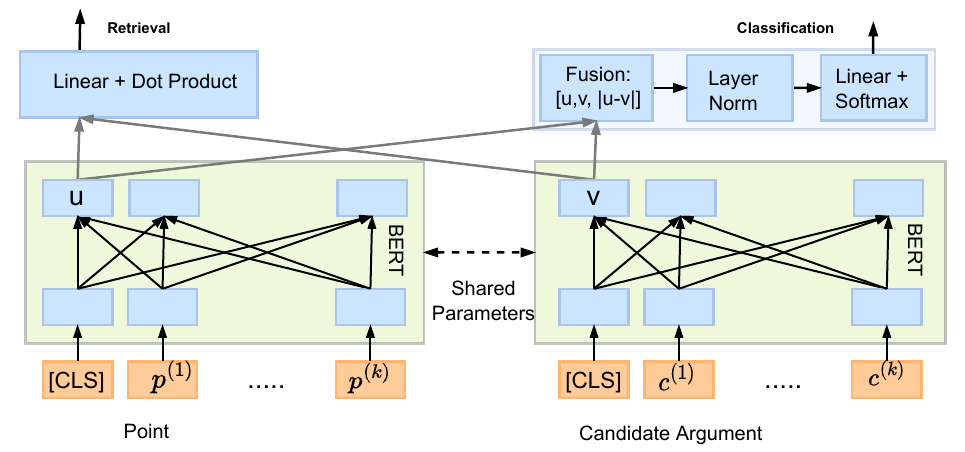}
 \caption{Bipolar-encoder architecture four counter-argument retrieval for a given argument point. The architecture contains a retrieval head for measure the ``similarity'' and a classification head for infering the ``dissimilarity''.}
 \label{fig:bipolar}
\end{figure*}

\paragraph{Efficiency} During evaluation, Bi-encoder is much more efficient than Cross-encoder. This is because Bi-encoder allows precomputation of embeddings for all the candidates. For a new argument point, the online calculation includes one embedding computation and many efficient dot-product operations. When being used with a nearest neighbor library such as FAISS \cite{johnson2021billion-faiss}, Bi-encoder can scale to millions of candidates. Cross-encoder, on the other hand, needs to concatenate a point and a candidate to calculate an embedding, and such expensive operation needs to be done for all the candidates during evaluation.

\paragraph{Effectiveness} Using early concatenation, Cross-encoder allows richer interaction between the input texts, thus, leading to better performance than Bi-encoders in many text matching tasks. To effectively use a Cross-encoder, one can employ an efficient retrieval model to select a small set of candidates, then use Cross-encoder for reranking. 

\section{Bipolar-Encoder}\label{sec:bipolar-encoder}
Being constructed for general text matching tasks,  Bi- and Cross-encoders are mostly optimized for semantic similarity. This paper proposes Bipolar-encoder, a new architecture that models simultaneous similarity and dissimilarity. This new architecture leads to performance gains over both Bi- and Cross-encoder while maintaining the level of efficiency closer to that of Bi-encoder.

\subsection{General Architecture}
 {\modelname} is demonstrated in Figure \ref{fig:bipolar}, which contains 3 main components: 1) the Encoding that converts texts into vectors; 2) the Retrieval Head that calculates the similarity; and 3) the Classification Head that integrates the dissimilarity cues.

\paragraph{Encoding} Like Bi-encoder, the argument point and the candidate are encoded by passing the corresponding sequences to BERT encoders and taking the output representations of the [CLS] tokens. Unlike Bi-encoder, however, the two encoders share parameters. This is because our preliminary results show that sharing parameters leads to more stable and faster convergence. The base embeddings for the argument point and the candidate are $\textbf{u}=BERT(p)$, $\textbf{v}=BERT(c)$, respectively. 

\paragraph{Retrieval Head} The upper-left part in our architecture (see Figure \ref{fig:bipolar}) corresponds to the retrieval component. Here, the base embeddings ($\textbf{u},\textbf{v}$) are first passed through a linear layer, which is to add more flexibility to model semantic similarity. The outputs of the linear layer can be considered as the \textbf{retrieval embeddings} $E_R$ and be used to calculate the retrieval similarity score:
$$\textbf{u}^{(1)}=\text{linear}(\textbf{u}); \textbf{v}^{(1)}=\text{linear}(\textbf{v})$$
$$f^{(ret)}(p,c)=\text{dot}(\textbf{u}^{(1)}, \textbf{v}^{(1)})$$

\paragraph{Classification Head} The upper-right part in our architecture (see Figure \ref{fig:bipolar}) corresponds to the classification component, which includes a fusion layer, a layer norm, a linear layer and softmax. 
$$\textbf{x}=\text{layer-norm}(\text{concat}(\textbf{u},\textbf{v}, |\textbf{u}-\textbf{v}|))$$
$$f^{(cls)}(p,c)=\hat{y}=\text{softmax}(\text{linear}(\textbf{x}))$$
The fusion layer concatenates $\textbf{u}$,  $\textbf{v}$, and $|\textbf{u}-\textbf{v}|$ which is the vector of absolute differences between two base embeddings $\textbf{u}$ and $\textbf{v}$. Intuitively, the difference vector $|\textbf{u}-\textbf{v}|$ helps modeling the dissimilarity between the input texts. Here, layer norm is added for faster training and better generalization. Finally, linear and softmax are exploited for classification like Cross-encoder.

\subsection{Training and Inference}
\paragraph{Training} Following the pairwise approach, our model is trained by simultaneously optimizing two losses, the triplet loss (for retrieval), and the cross-entropy loss (for classification), on the training set $\mathcal{D}^{(pair)}=\{\langle p_j,c^+_j, c^-_j \rangle\}_{j=1}^M$. Here, the combined loss for a triple $(p_j,c^+_j, c^-_j)$ is obtained by:
$$\mathcal{L}^{(j)}_{TL}(p_j,c^+_j,c^-_j)+\mathcal{L}^{j}_{CE}(p_j,c^+_j)+\mathcal{L}^{j}_{CE}(p_j,c^-_j)$$
where the cross-entropy loss and triplet loss are calculated as in Section \ref{sec:neural-models}

\paragraph{Inference} Bipolar-encoder follows the retrieve-and-rerank approach during inference. Specifically, Bipolar-encoder first precomputes base and retrieval embeddings of all candidates in the collection. Given an argument point, the retrieval head is used to obtain a shortlist of candidates. We then rerank the candidates based on the classification, which makes uses of the precomputed base embeddings, and thus more efficient than Cross-encoder. 



\section{Experiments}\label{sec:experiments}
\subsection{Counterargs-18 Dataset}

\citet{wachsmuth2018retrieval} introduced {\dsname} corpus containing argument pairs from $1069$ debates. The corpus covers wide range of $15$ themes such as culture, religion, society and sport. The total number of arguments in the collection is $12506$. Each argument is paired with one correct counter-argument that directly address the aspect of the input argument from the opposite stance. The dataset is split into a training set, containing of the first 60\% of all debates of each theme, a validation set and a test set each covering 20\%. 

\subsection{Traditional Scoring Models}

\paragraph{Baselines and Evaluation} We compare Simple-SD and other traditional LTR methods on 4 retrieval settings (\textbf{sdoc, sda, epc, epa}) (see Section \ref{sec:background} for more details). For fair comparison, we consider two settings: 1) Logit-part and XGBoost-part which use 8 features from Table \ref{tab:features} like Simple-SD; and 2) Logit-full and XGBoost-full which exploit all the features from Table \ref{tab:features}. All approaches are evaluated with accuracy@1 since there is only one correct counter for each argument point. 

\paragraph{Results and Discussion} Table \ref{tab:simple-baselines} shows the experimental results of traditional baselines, from which several observations can be drawn. First, XGBoost methods are generally better than other traditional baselines, with the best performance can be achieved with XGBoost-full. Second, logistic regression baselines can only achieve comparable results with Simple-SD (our implementation), and the inclusion of more features in Logit-full does not show more advantages.  Finally, the \textbf{epa} setting (topic/stance unknown) is very challenging, even XGBoost-full can obtain only a marginal gain. 

It should be noted that these traditional baselines are very inefficient at inference time. This is because the feature extraction step requires online calculation of different similarity scores for all the candidates, in which some are expensive to compute such as the Earth-Mover distance. One may apply approximate search to reduce the number of candidates but then we need to sacrifice the already-poor performance.


\begin{table}
    \centering
    \begin{tabular}{lcccc}
        \toprule
         \multirow{2}{*}{\small{\textbf{Approach}}} & \multicolumn{4}{c}{\small{\textbf{Accuracy@1}}} \\
         \cline{2-5}
          & \small{\textbf{Sdoc}} & \small{\textbf{Sda}} & \small{\textbf{Epc}} & \small{\textbf{Epa}} \\
        \midrule
        Simple-SD & 74.5 & 44.1 & 50.0 & 32.4 \\
        Simple-SD \dag & 72.4 & 41.8 & 44.7 & 31.0 \\
        Logit-part & 73.7 & 42.3 & 45.4 & 30.0 \\ 
        Logit-full & 73.6 & 43.0 & 44.9 & 31.0 \\
        XGBoost-part & 77.6 & 43.4 & 44.2 & 31.3 \\
        XGBoost-full & \textbf{79.7} & \textbf{47.5} & \textbf{47.3} & \textbf{33.7} \\
        \toprule
    \end{tabular}
    \caption{Results of traditional baselines on sdoc, sda, epc and epa settings. Here, Simple-SD{\dag} is our reimplementation, and Logit is short for Logistic Regression.}
    \label{tab:simple-baselines}
\end{table}

\subsection{Neural Scoring Methods}

\paragraph{Baselines and Evaluation} We compare \textbf{Bipolar-encoder} with multiple baselines: 1) the retrieval approach including \textbf{Bi-encoder} and \textbf{Unipolar-ret}, the simplified variant of Bipolar-encoder with only the retrieval head. The main difference between Bi-encoder and Unipolar-ret is that the former one uses two separate BERT for encoding the point and the candidate, while the later shares one BERT. 2) the retrieve-then-rerank approach, which couples an efficient retrieval model such as \textbf{BM25} and \textbf{Bi-encoder} with a classification model such as \textbf{Cross-encoder} or \textbf{Unipolar-cls}. Here, Unipolar-cls is the simplified variant of Bipolar-encoder with only the classification head. 

For ablation study, we consider two more variants of Bipolar-encoder: 1) \textbf{Bipolar-encoder} (w/o joint): instead of joint-training and get one model with two heads (retrieval, classification), we train Unipolar-ret and Unipolar-cls separately; then use Unipolar-ret for retrieval and Unipolar-cls for classification; 2) \textbf{Bipolar-encoder} (w/o $|\textbf{u}-\textbf{v}|$): we train \textbf{Bipolar-encoder} without the difference (dissimilarity measure) $|\textbf{u}-\textbf{v}|$ in the fusion layer of the classification head. 

All methods are evaluated using $@1/K$, which is the accuracy@1 obtained by (re)-ranking $K$ retrieved candidates. For retrieval models without a reranking component, $@1/K$ are the same for all $K$, which is the accuracy@1 of the original rank obtained by the retrieval model.

\begin{table}
    \centering
    \begin{tabular}{llccc}
        \toprule
         \multirow{2}{*}{\small{\textbf{Retrieve}}} & \multirow{2}{*}{\small{\textbf{Rerank}}} & \multicolumn{3}{c}{\small{\textbf{Epa Setting}}} \\
         \cline{3-5}
          & & \small{\textbf{@1/10}} & \small{\textbf{@1/30}} & \small{\textbf{@1/50}} \\
        Bi-enc & - & 14.88 & - & -\\
        Unipo-ret & - & 30.04  & - & - \\ 
        BM25 & Cross-enc & 8.76 & 15.57 & 15.57 \\
        BM25 & Unipo-cls & 1.28 & 2.16 & 2.73\\
        Bi-enc & Cross-enc &  6.22 & 2.44 & 1.79 \\
        \midrule
        \multicolumn{2}{l}{Bipolar-encoder} & \textbf{48.86} & \textbf{49.04} & \textbf{48.89}  \\ 
        \multicolumn{2}{l}{- w/o |u-v|} &  13.40 & 5.48 & 1.24\\
        \multicolumn{2}{p{3cm}}{- w/o joint}  & 43.63 & 42.02 & 41.74 \\
        \toprule
    \end{tabular}
    \caption{Results of neural scoring models on the \textbf{epa} setting. Here, Bi-enc and Cross-enc are for Bi- Cross-encoder, and Unipo- is short for Unipolar-.}
    \label{tab:neural-baselines}
\end{table}

\paragraph{Results and Discussion} Table \ref{tab:neural-baselines} summarizes the results of neural scoring models when topics and stances are unknown (\textbf{epc} setting). Several observations can be drawn as follows: 

Retrieval models can only obtain limited success, implying that using only similarity is not effective. Specifically, Unipolar-ret achieves accuracy@1 of 30.04\%, which is at the same level with traditional methods. Bi-encoder can reach accuracy@1 of 14.88\%, not performs well. Although our retrieval variant (Unipolar-ret) is inferior to Bi-encoder, Unipolar-ret is much easier to converge during training compared to the later one (Bi-encoder). It should also be noted that Bi-encoder and Unipolar-ret are more efficient than traditional methods.

Both similarity measured by the retrieval and dissimilarity captured by the reranker are important for counter-argument retrieval. Using the same retrieval model but BM25+Unipolar-cls performs consistently better than BM25+Cross-encoder for all @1/K (K=10,50,100). This implies that the proposed architecture for Unipolar-cls is better than Cross-encoder.  Our model (Bipolar-encoder), which explicitly makes use of dissimilarity, outperforms the best neural baseline (Bi-enc+Cross-enc). Particularly, Bipolar-encoder can achieve  accuracy@1 of $49.04$ when its retrieval head of the model selects 30 candidates for its classification head to rerank.

Ablation study confirms the importance of dissimilarity in Bipolar-encoder, but does not show clear advantages of joint training. Without the difference $|\textbf{u}-\textbf{v}|$, the performance of Bipolar-encoder drops significantly (from @1/10 of 48.86 to @1/10 of 13.40). On the other hand, we observe inconsistent results when comparing Bipolar-encoder and Bipolar-encoder (w/o joint) on different @1/K. One possible explanation is that joint training in Bipolar-encoder helps improve the retrieval model but worsens the classification model. When only 10 candidates are considered, the gain in the retrieval model outweighs the loss in the classification, leading to the superior of Bipolar-encoder. When more candidates are selected, the role of the classification model becomes more significant than the retrieval model, making Bipolar-encoder (w/o joint) slightly better than Bipolar-encoder. We expect to investigate this issue further in the future work. 





\section{Conclusion}
This paper evaluates different approaches to finding best counter-argument. Our study shows that traditional learning-to-rank models based on Logistic Regression and XGBoost are both inefficient and ineffective, particularly when the information on topics and stances is not available. On the other hand, we observe significant improvements with neural scoring methods following retrieve-then-rerank approach. Since only a shortlist of candidates are obtained for reranking, such methods are also much more efficient than the traditional methods. In addition, we propose a novel neural architecture, Bipolar-encoder that explicitly models similarity and dissimilarity, leading to further performance gain. This confirms that both similarity and dissimilarity are important for our task.

\section*{Limitations}
The task of best counter-argument retrieval is very challenging. Although we can avoid the difficult tasks of topics and stance detection, this study assumes that argumentative units (argument points) needs to be recognized first. In addition, this study not yet considers argument qualities such as logical, persuasion, and so on, which are very important factors in assessing arguments. Last but not least, although our model can find best counter-arguments more effectively, it can not provide insights on how the counter-argument attacks the argument point. Knowing different kinds of attack would be of great interest for many applications such as essay scoring, AI in judicial systems.

\section*{Ethics Statement}

Like many other AI applications, a system of best counter-argument retrieval can be used towards both beneficial and malicious ends. It can be used in the fight against fake news or disinformation. It can also be used to find fake news or conspiracy theories which counters legitimate arguments. As a result, further investigation needs to be done to integrate trustworthy qualities into counter-argument retrieval. Another ethical issue that require our attention is the problem of biases, which may come from the training data (argumentative collections), the pre-trained model (BERT), or the training algorithm (negative sampling strategies). In practice, a counter-argument retrieval should be tested thoroughly to remove as much biases as possible.


\section*{Acknowledgements}

\bibliography{anthology,revisiting}

\begin{thebibliography}{41}
\expandafter\ifx\csname natexlab\endcsname\relax\def\natexlab#1{#1}\fi

\bibitem[{Al~Khatib et~al.(2021)Al~Khatib, Trautner, Wachsmuth, Hou, and
  Stein}]{al-khatib-etal-2021-employing}
Khalid Al~Khatib, Lukas Trautner, Henning Wachsmuth, Yufang Hou, and Benno
  Stein. 2021.
\newblock \href {https://doi.org/10.18653/v1/2021.acl-long.366} {Employing
  argumentation knowledge graphs for neural argument generation}.
\newblock In \emph{Proceedings of the 59th Annual Meeting of the Association
  for Computational Linguistics and the 11th International Joint Conference on
  Natural Language Processing (Volume 1: Long Papers)}, pages 4744--4754,
  Online. Association for Computational Linguistics.

\bibitem[{ALDayel and Magdy(2021)}]{ALDayel2021stance}
Abeer ALDayel and Walid Magdy. 2021.
\newblock \href {https://doi.org/https://doi.org/10.1016/j.ipm.2021.102597}
  {Stance detection on social media: State of the art and trends}.
\newblock \emph{Information Processing \& Management}, 58(4):102597.

\bibitem[{Alshomary et~al.(2021)Alshomary, Syed, Dhar, Potthast, and
  Wachsmuth}]{alshomary-etal-2021-counter}
Milad Alshomary, Shahbaz Syed, Arkajit Dhar, Martin Potthast, and Henning
  Wachsmuth. 2021.
\newblock \href {https://doi.org/10.18653/v1/2021.findings-acl.159}
  {Counter-argument generation by attacking weak premises}.
\newblock In \emph{Findings of the Association for Computational Linguistics:
  ACL-IJCNLP 2021}, pages 1816--1827, Online. Association for Computational
  Linguistics.

\bibitem[{Bao et~al.(2021)Bao, Fan, Wu, Dang, Du, and
  Xu}]{bao-etal-2021-neural}
Jianzhu Bao, Chuang Fan, Jipeng Wu, Yixue Dang, Jiachen Du, and Ruifeng Xu.
  2021.
\newblock \href {https://doi.org/10.18653/v1/2021.acl-long.497} {A neural
  transition-based model for argumentation mining}.
\newblock In \emph{Proceedings of the 59th Annual Meeting of the Association
  for Computational Linguistics and the 11th International Joint Conference on
  Natural Language Processing (Volume 1: Long Papers)}, pages 6354--6364,
  Online. Association for Computational Linguistics.

\bibitem[{Barrow et~al.(2021)Barrow, Jain, Lipka, Dernoncourt, Morariu,
  Manjunatha, Oard, Resnik, and Wachsmuth}]{barrow-etal-2021-syntopical}
Joe Barrow, Rajiv Jain, Nedim Lipka, Franck Dernoncourt, Vlad Morariu, Varun
  Manjunatha, Douglas Oard, Philip Resnik, and Henning Wachsmuth. 2021.
\newblock \href {https://doi.org/10.18653/v1/2021.acl-long.126} {Syntopical
  graphs for computational argumentation tasks}.
\newblock In \emph{Proceedings of the 59th Annual Meeting of the Association
  for Computational Linguistics and the 11th International Joint Conference on
  Natural Language Processing (Volume 1: Long Papers)}, pages 1583--1595,
  Online. Association for Computational Linguistics.

\bibitem[{Cabrio and Villata(2018)}]{cabrio-serena-2018five-years}
Elena Cabrio and Serena Villata. 2018.
\newblock Five years of argument mining: A data-driven analysis.
\newblock In \emph{Proceedings of the 27th International Joint Conference on
  Artificial Intelligence}, IJCAI'18, page 5427–5433. AAAI Press.

\bibitem[{Cha(2007)}]{cha2007comprehensive}
Sung-Hyuk Cha. 2007.
\newblock Comprehensive survey on distance/similarity measures between
  probability density functions.
\newblock \emph{City}, 1(2):1.

\bibitem[{Chen and Guestrin(2016)}]{chen2016xgboost}
Tianqi Chen and Carlos Guestrin. 2016.
\newblock Xgboost: A scalable tree boosting system.
\newblock In \emph{Proceedings of the 22nd acm sigkdd international conference
  on knowledge discovery and data mining}, pages 785--794.

\bibitem[{Cheng et~al.(2021)Cheng, Wu, Bing, and Si}]{cheng-etal-2021-argument}
Liying Cheng, Tianyu Wu, Lidong Bing, and Luo Si. 2021.
\newblock \href {https://doi.org/10.18653/v1/2021.acl-long.496} {Argument pair
  extraction via attention-guided multi-layer multi-cross encoding}.
\newblock In \emph{Proceedings of the 59th Annual Meeting of the Association
  for Computational Linguistics and the 11th International Joint Conference on
  Natural Language Processing (Volume 1: Long Papers)}, pages 6341--6353,
  Online. Association for Computational Linguistics.

\bibitem[{Cocarascu and Toni(2017)}]{cocarascu-toni-2017-identifying}
Oana Cocarascu and Francesca Toni. 2017.
\newblock \href {https://doi.org/10.18653/v1/D17-1144} {Identifying attack and
  support argumentative relations using deep learning}.
\newblock In \emph{Proceedings of the 2017 Conference on Empirical Methods in
  Natural Language Processing}, pages 1374--1379, Copenhagen, Denmark.
  Association for Computational Linguistics.

\bibitem[{Damer(2012)}]{damer2012attacking}
T~Edward Damer. 2012.
\newblock \emph{Attacking faulty reasoning}.
\newblock Cengage Learning.

\bibitem[{Devlin et~al.(2019)Devlin, Chang, Lee, and
  Toutanova}]{devlin-etal-2019-bert}
Jacob Devlin, Ming-Wei Chang, Kenton Lee, and Kristina Toutanova. 2019.
\newblock \href {https://doi.org/10.18653/v1/N19-1423} {{BERT}: Pre-training of
  deep bidirectional transformers for language understanding}.
\newblock In \emph{Proceedings of the 2019 Conference of the North {A}merican
  Chapter of the Association for Computational Linguistics: Human Language
  Technologies, Volume 1 (Long and Short Papers)}, pages 4171--4186,
  Minneapolis, Minnesota. Association for Computational Linguistics.

\bibitem[{Dumani and Schenkel(2020)}]{lorik2020quality-aware}
Lorik Dumani and Ralf Schenkel. 2020.
\newblock Quality-aware ranking of arguments.
\newblock In \emph{Proceedings of the 29th ACM International Conference on
  Information \&amp; Knowledge Management}, CIKM '20, page 335–344, New York,
  NY, USA. Association for Computing Machinery.

\bibitem[{Emelin et~al.(2021)Emelin, Le~Bras, Hwang, Forbes, and
  Choi}]{emelin-etal-2021-moral}
Denis Emelin, Ronan Le~Bras, Jena~D. Hwang, Maxwell Forbes, and Yejin Choi.
  2021.
\newblock \href {https://doi.org/10.18653/v1/2021.emnlp-main.54} {Moral
  stories: Situated reasoning about norms, intents, actions, and their
  consequences}.
\newblock In \emph{Proceedings of the 2021 Conference on Empirical Methods in
  Natural Language Processing}, pages 698--718, Online and Punta Cana,
  Dominican Republic. Association for Computational Linguistics.

\bibitem[{Feng and Hirst(2011)}]{feng-hirst-2011-classifying}
Vanessa~Wei Feng and Graeme Hirst. 2011.
\newblock \href {https://aclanthology.org/P11-1099} {Classifying arguments by
  scheme}.
\newblock In \emph{Proceedings of the 49th Annual Meeting of the Association
  for Computational Linguistics: Human Language Technologies}, pages 987--996,
  Portland, Oregon, USA. Association for Computational Linguistics.

\bibitem[{Hua et~al.(2019)Hua, Hu, and Wang}]{hua-etal-2019argument-generation}
Xinyu Hua, Zhe Hu, and Lu~Wang. 2019.
\newblock Argument generation with retrieval, planning, and realization.
\newblock In \emph{Proceedings of the 57th Annual Meeting of the Association
  for Computational Linguistics}, pages 2661--2672, Florence, Italy.

\bibitem[{Hua and Wang(2018)}]{hua-wang-2018neural}
Xinyu Hua and Lu~Wang. 2018.
\newblock Neural argument generation augmented with externally retrieved
  evidence.
\newblock In \emph{Proceedings of the 56th Annual Meeting of the Association
  for Computational Linguistics (Volume 1: Long Papers)}, pages 219--230,
  Melbourne, Australia.

\bibitem[{Humeau et~al.(2019)Humeau, Shuster, Lachaux, and
  Weston}]{humeau2019poly}
Samuel Humeau, Kurt Shuster, Marie-Anne Lachaux, and Jason Weston. 2019.
\newblock Poly-encoders: Architectures and pre-training strategies for fast and
  accurate multi-sentence scoring.
\newblock In \emph{International Conference on Learning Representations}.

\bibitem[{Jo et~al.(2021)Jo, Yoo, Bak, Oh, Reed, and
  Hovy}]{jo-etal-2021-knowledge-enhanced}
Yohan Jo, Haneul Yoo, JinYeong Bak, Alice Oh, Chris Reed, and Eduard Hovy.
  2021.
\newblock \href {https://doi.org/10.18653/v1/2021.findings-emnlp.264}
  {Knowledge-enhanced evidence retrieval for counterargument generation}.
\newblock In \emph{Findings of the Association for Computational Linguistics:
  EMNLP 2021}, pages 3074--3094, Punta Cana, Dominican Republic. Association
  for Computational Linguistics.

\bibitem[{Johnson et~al.(2021)Johnson, Douze, and
  Jégou}]{johnson2021billion-faiss}
Jeff Johnson, Matthijs Douze, and Hervé Jégou. 2021.
\newblock Billion-scale similarity search with gpus.
\newblock \emph{IEEE Transactions on Big Data}, 7(3):535--547.

\bibitem[{Karpukhin et~al.(2020)Karpukhin, Oguz, Min, Lewis, Wu, Edunov, Chen,
  and Yih}]{karpukhin-etal-2020-dense}
Vladimir Karpukhin, Barlas Oguz, Sewon Min, Patrick Lewis, Ledell Wu, Sergey
  Edunov, Danqi Chen, and Wen-tau Yih. 2020.
\newblock \href {https://doi.org/10.18653/v1/2020.emnlp-main.550} {Dense
  passage retrieval for open-domain question answering}.
\newblock In \emph{Proceedings of the 2020 Conference on Empirical Methods in
  Natural Language Processing (EMNLP)}, pages 6769--6781, Online. Association
  for Computational Linguistics.

\bibitem[{Kusner et~al.(2015)Kusner, Sun, Kolkin, and
  Weinberger}]{kusner2015word}
Matt Kusner, Yu~Sun, Nicholas Kolkin, and Kilian Weinberger. 2015.
\newblock From word embeddings to document distances.
\newblock In \emph{International conference on machine learning}, pages
  957--966. PMLR.

\bibitem[{Lawrence and Reed(2019)}]{lawrence2019argument}
John Lawrence and Chris Reed. 2019.
\newblock Argument mining: A survey.
\newblock \emph{Computational Linguistics}, 45(4):765--818.

\bibitem[{Lazer et~al.(2018)Lazer, Baum, Benkler, Berinsky, Greenhill, Menczer,
  Metzger, Nyhan, Pennycook, Rothschild et~al.}]{lazer2018science}
David~MJ Lazer, Matthew~A Baum, Yochai Benkler, Adam~J Berinsky, Kelly~M
  Greenhill, Filippo Menczer, Miriam~J Metzger, Brendan Nyhan, Gordon
  Pennycook, David Rothschild, et~al. 2018.
\newblock The science of fake news.
\newblock \emph{Science}, 359(6380):1094--1096.

\bibitem[{Le et~al.(2018)Le, Nguyen, and Nguyen}]{le2018dave}
Dieu-Thu Le, Cam-Tu Nguyen, and Kim~Anh Nguyen. 2018.
\newblock Dave the debater: a retrieval-based and generative argumentative
  dialogue agent.
\newblock In \emph{Proceedings of the 5th Workshop on Argument Mining}, pages
  121--130.

\bibitem[{Li(2011)}]{li2011short}
Hang Li. 2011.
\newblock A short introduction to learning to rank.
\newblock \emph{IEICE TRANSACTIONS on Information and Systems},
  94(10):1854--1862.

\bibitem[{Orbach et~al.(2020)Orbach, Bilu, Toledo, Lahav, Jacovi, Aharonov, and
  Slonim}]{orbach2020echo}
Matan Orbach, Yonatan Bilu, Assaf Toledo, Dan Lahav, Michal Jacovi, Ranit
  Aharonov, and Noam Slonim. 2020.
\newblock Out of the echo chamber: {D}etecting countering debate speeches.
\newblock In \emph{Proceedings of the 58th Annual Meeting of the Association
  for Computational Linguistics}, pages 7073--7086.

\bibitem[{Potthast et~al.(2019)Potthast, Gienapp, Euchner, Heilenk{\"o}tter,
  Weidmann, Wachsmuth, Stein, and Hagen}]{potthast2019argument}
Martin Potthast, Lukas Gienapp, Florian Euchner, Nick Heilenk{\"o}tter, Nico
  Weidmann, Henning Wachsmuth, Benno Stein, and Matthias Hagen. 2019.
\newblock Argument search: assessing argument relevance.
\newblock In \emph{Proceedings of the 42nd International ACM SIGIR Conference
  on Research and Development in Information Retrieval}, pages 1117--1120.

\bibitem[{Qu et~al.(2021)Qu, Ding, Liu, Liu, Ren, Zhao, Dong, Wu, and
  Wang}]{qu-etal-2021-rocketqa}
Yingqi Qu, Yuchen Ding, Jing Liu, Kai Liu, Ruiyang Ren, Wayne~Xin Zhao, Daxiang
  Dong, Hua Wu, and Haifeng Wang. 2021.
\newblock \href {https://doi.org/10.18653/v1/2021.naacl-main.466}
  {{R}ocket{QA}: An optimized training approach to dense passage retrieval for
  open-domain question answering}.
\newblock In \emph{Proceedings of the 2021 Conference of the North American
  Chapter of the Association for Computational Linguistics: Human Language
  Technologies}, pages 5835--5847, Online. Association for Computational
  Linguistics.

\bibitem[{Robertson et~al.(2009)Robertson, Zaragoza
  et~al.}]{robertson2009probabilistic}
Stephen Robertson, Hugo Zaragoza, et~al. 2009.
\newblock The probabilistic relevance framework: Bm25 and beyond.
\newblock \emph{Foundations and Trends{\textregistered} in Information
  Retrieval}, 3(4):333--389.

\bibitem[{Slonim et~al.(2021)Slonim, Bilu, Alzate, Bar-Haim, Bogin, Bonin,
  Choshen, Cohen-Karlik, Dankin, Edelstein et~al.}]{slonim2021autonomous}
Noam Slonim, Yonatan Bilu, Carlos Alzate, Roy Bar-Haim, Ben Bogin, Francesca
  Bonin, Leshem Choshen, Edo Cohen-Karlik, Lena Dankin, Lilach Edelstein,
  et~al. 2021.
\newblock An autonomous debating system.
\newblock \emph{Nature}, 591(7850):379--384.

\bibitem[{Speer et~al.(2017)Speer, Chin, and Havasi}]{speer2017conceptnet}
Robyn Speer, Joshua Chin, and Catherine Havasi. 2017.
\newblock Conceptnet 5.5: An open multilingual graph of general knowledge.
\newblock In \emph{Thirty-first AAAI conference on artificial intelligence}.

\bibitem[{Sun et~al.(2018)Sun, Wang, Zhu, and Zhou}]{sun-etal-2018-stance}
Qingying Sun, Zhongqing Wang, Qiaoming Zhu, and Guodong Zhou. 2018.
\newblock \href {https://aclanthology.org/C18-1203} {Stance detection with
  hierarchical attention network}.
\newblock In \emph{Proceedings of the 27th International Conference on
  Computational Linguistics}, pages 2399--2409, Santa Fe, New Mexico, USA.
  Association for Computational Linguistics.

\bibitem[{Tan et~al.(2016)Tan, Niculae, Danescu-Niculescu-Mizil, and
  Lee}]{tan2016winning}
Chenhao Tan, Vlad Niculae, Cristian Danescu-Niculescu-Mizil, and Lillian Lee.
  2016.
\newblock Winning arguments: Interaction dynamics and persuasion strategies in
  good-faith online discussions.
\newblock In \emph{Proceedings of the 25th international conference on world
  wide web}, pages 613--624.

\bibitem[{Vassiliades et~al.(2021)Vassiliades, Bassiliades, and
  Patkos}]{vassiliades2021argumentation}
Alexandros Vassiliades, Nick Bassiliades, and Theodore Patkos. 2021.
\newblock Argumentation and explainable artificial intelligence: a survey.
\newblock \emph{The Knowledge Engineering Review}, 36.

\bibitem[{Vecchi et~al.(2021)Vecchi, Falk, Jundi, and
  Lapesa}]{vecchi-etal-2021-towards}
Eva~Maria Vecchi, Neele Falk, Iman Jundi, and Gabriella Lapesa. 2021.
\newblock \href {https://doi.org/10.18653/v1/2021.acl-long.107} {Towards
  argument mining for social good: A survey}.
\newblock In \emph{Proceedings of the 59th Annual Meeting of the Association
  for Computational Linguistics and the 11th International Joint Conference on
  Natural Language Processing (Volume 1: Long Papers)}, pages 1338--1352,
  Online. Association for Computational Linguistics.

\bibitem[{Wachsmuth et~al.(2018)Wachsmuth, Syed, and
  Stein}]{wachsmuth2018retrieval}
Henning Wachsmuth, Shahbaz Syed, and Benno Stein. 2018.
\newblock Retrieval of the best counterargument without prior topic knowledge.
\newblock In \emph{Proceedings of the 56th Annual Meeting of the Association
  for Computational Linguistics (Volume 1: Long Papers)}, pages 241--251.

\bibitem[{Walton(2009)}]{walton2009argumentation-theory}
Douglas Walton. 2009.
\newblock \emph{Argumentation Theory: A Very Short Introduction}, pages 1--22.
  Springer US, Boston, MA.

\bibitem[{Wambsganss et~al.(2021)Wambsganss, Niklaus, S{\"o}llner, Handschuh,
  and Leimeister}]{wambsganss-etal-2021-supporting}
Thiemo Wambsganss, Christina Niklaus, Matthias S{\"o}llner, Siegfried
  Handschuh, and Jan~Marco Leimeister. 2021.
\newblock \href {https://doi.org/10.18653/v1/2021.acl-long.314} {Supporting
  cognitive and emotional empathic writing of students}.
\newblock In \emph{Proceedings of the 59th Annual Meeting of the Association
  for Computational Linguistics and the 11th International Joint Conference on
  Natural Language Processing (Volume 1: Long Papers)}, pages 4063--4077,
  Online. Association for Computational Linguistics.

\bibitem[{Yuan(2021)}]{yuan-2021-interactive}
Xingdi Yuan. 2021.
\newblock \href {https://doi.org/10.18653/v1/2021.emnlp-main.540} {Interactive
  machine comprehension with dynamic knowledge graphs}.
\newblock In \emph{Proceedings of the 2021 Conference on Empirical Methods in
  Natural Language Processing}, pages 6734--6750, Online and Punta Cana,
  Dominican Republic. Association for Computational Linguistics.

\bibitem[{Zhao et~al.(2021)Zhao, Durmus, Zhang, and
  Cardie}]{zhao-etal-2021-leveraging}
Xinran Zhao, Esin Durmus, Hongming Zhang, and Claire Cardie. 2021.
\newblock \href {https://doi.org/10.18653/v1/2021.findings-acl.386} {Leveraging
  topic relatedness for argument persuasion}.
\newblock In \emph{Findings of the Association for Computational Linguistics:
  ACL-IJCNLP 2021}, pages 4401--4407, Online. Association for Computational
  Linguistics.

\end{thebibliography}
\bibliographystyle{acl_natbib}

\appendix

\section{CounterArgs-18 Dataset}
\label{sec:apppendix-dataset}
\begin{figure}
    \centering
    \includegraphics[width=0.49\textwidth]{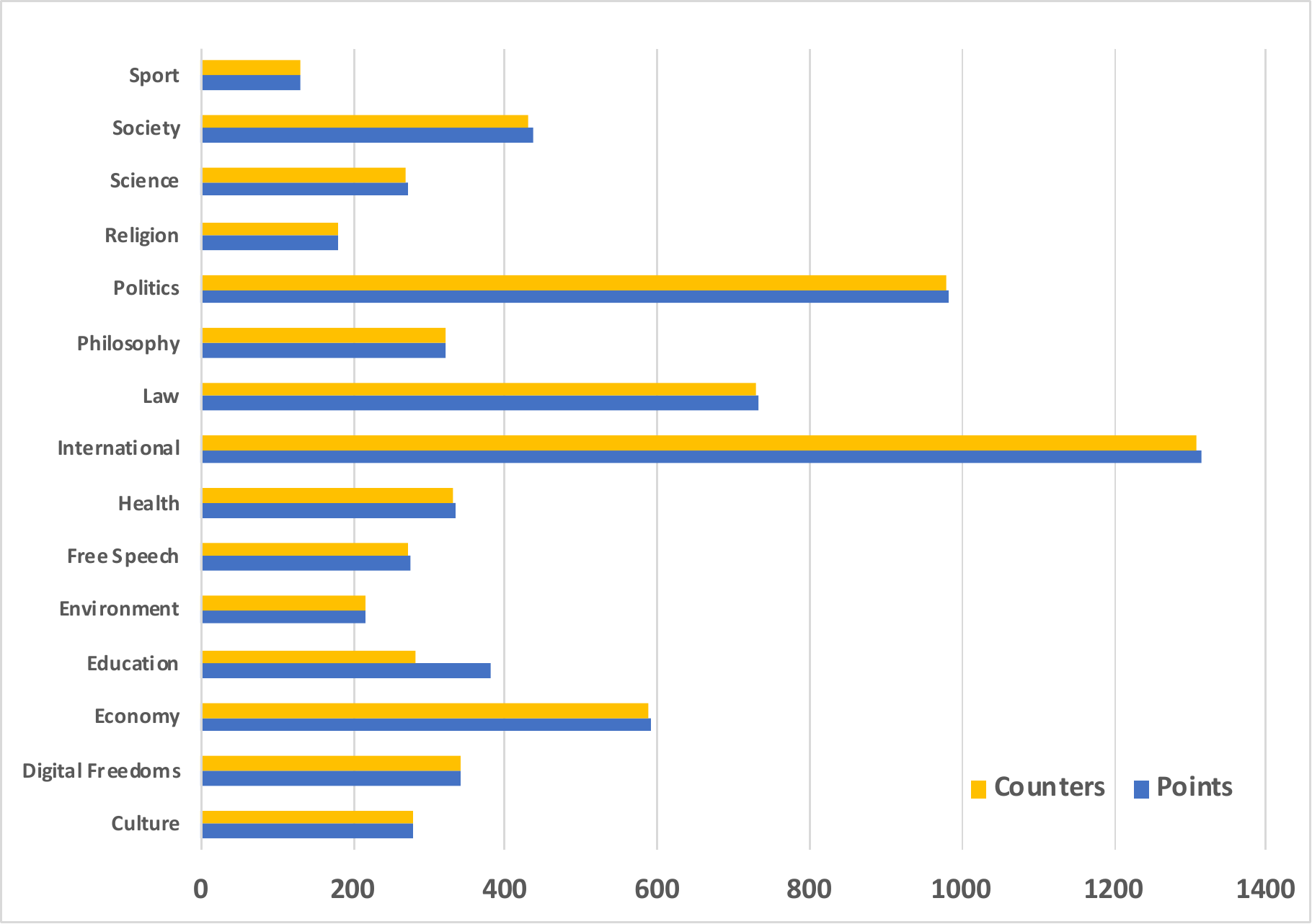}
    \caption{The number of Points \& Counters by Domain}
    \label{fig:nums-by-domain}
\end{figure}
The number of points \& counters by domain is shown in the figure \ref{fig:nums-by-domain}. The CounterArgs-18 Dataset consists of 1069 debates with 6753 points that have a counter. The mean length of points is 196.3 words, whereas counters span only 129.6 words. After preprocessing, we get our training set 8130 arguments and validation set 2574 arguments and test set 2802 arguments.

\section{Experimental Details}

\subsection{Traditional Scoring Methods}

\paragraph{XGBoost} 
 We used the XGBoost package  in Python. For data preprocessing, we used StandardScaler from sklearn package.
 Hyper-parameters used by XGBoost as follows, other hyper-parameters were set as defaults: \\ ~\\
  max\_depth=8 \\
  learning\_rate=0.01 \\
  n\_estimators=1000 \\
  min\_child\_weight=5 \\
  max\_delta\_step=0 \\
  subsample=0.8  \\
  colsample\_bytree=0.7 \\
  reg\_alpha=0 \\
  reg\_lambda=0.4 \\
  scale\_pos\_weight=0.8 \\
  objective=binary:logistic \\
  eval\_metric=auc \\
  gamma=0 \\
  n\_jobs=20 \\

\paragraph{Logistic Regression} We used LogisticRegression from sklearn. For data preprocessing, we used StandardScaler from sklearn to transform data into normal distribution, all hyper-parameters in StandardScaler were set as defaults. The  class\_weight was set to be 'balanced', max\_iter of 1000 and C of 50, and other hyper-parameters were set as the default. \\ 

\subsection{Neural Scoring Methods}


We considers 5 neural scoring models including Bi-encoder, Cross-encoder, Uni-polar-ret, Unipolar-cls, and Bipolar-encoder. The details about the architectures of Bi-encoder, Cross-encoder, Bipolar-encoder are given in the main text. In the following sections, we details the architectures of Unipolar-ret and Unipolar-cls, which are the simplifed versions of Bipolar-encoder to contain only retrieval or classification head. All 5 models used BERT as encoders. Here, we used default bert-base-uncased\footnote{\url{https://huggingface.co/bert-base-uncased}} from huggingface, implemented in pytorch \footnote{\url{https://pytorch.org/}}. There are 12 BertLayers in the bert-base-uncased model and 12 attention heads in each BertLayer. It is about 110M parameters in the model. 

All 5 models were trained with 200 epoch. We trained all our models on one V100 GPU card. For training, each model took around 60 hours to train. For inference, except for Cross-encoder which is very time consuming, all the models can finish inference within half an hour.

There are two negative sampling strategies that we used for training neural scoring models. The first one is the simple random sampling, which choose negative samples for a given argument point randomly. The second one is ``increasing hard negative sampling''. Specifically, for every training epoch $i$, with the probability $p=min(i*\text{increase\_rate},1)$, we will perform hard sampling. With the probability $1-p$, we will perform random sampling.  By hard sampling, we mean to choose negative samples closest to the positive one as samples. Here, $increase\_rate$ was set to 0.02. We used BallTree\footnote{\url{https://scikit-learn.org/stable/modules/generated/sklearn.neighbors.BallTree.html}} from sklearn for nearest neighbor search. In other words, we start with easy samples (random sampling), and increasing the ratio of hard negative sampling overtime. 

\subsubsection{Bi-encoder} 

 
 We implemented Bi-encoder with Tritplet loss, which is also a part of  Pytorch package.  We used Adam optimizer to train, and chose the batchsize of 4 and the learning\_rate of 3e-6. For Triplet loss part, the margin was choosen the value of 1.0 and p of 2. For negative sampling, ``increasing hard negative sampling'' strategy was used, for Bi-encoder can not converge well with ``random sampling''.\\

\subsubsection{Cross-encoder} 

We used random negative sampling for every epoch. The Adam learning\_rate was set to be 3e-6 and other hyper-parameters were set as the default. The batchsize was chosen to be 4.\\

\subsubsection{Unipolar-ret}
\begin{figure}
  \centering
  \includegraphics[width=0.45\textwidth]{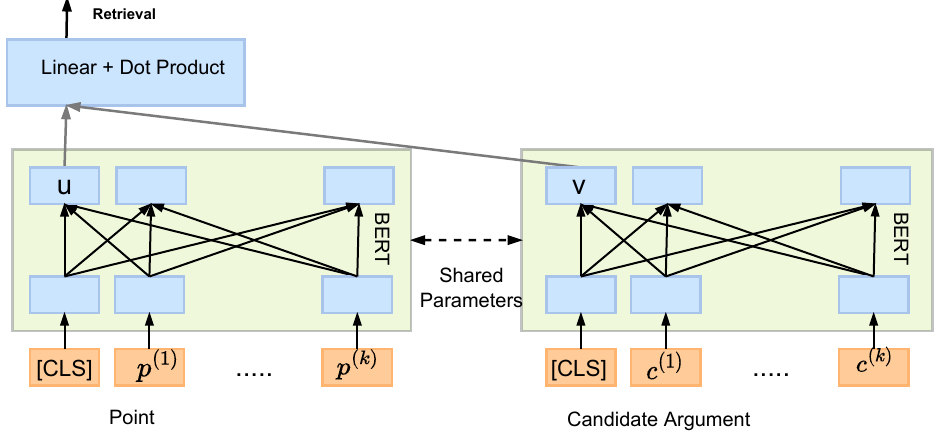}
 \caption{Unipolar-ret Architecture}
 \label{fig:unipolar-ret}
\end{figure}

Unipolar-ret is the simplified variant of Bipolar-encoder with only retrieval head. The architecture of Unipolar-ret is shown in Figure \ref{fig:unipolar-ret}. Unipolar-ret is trained based on Triplet loss on the dataset $\mathcal{D}^{pair}=\{\langle p_j,c^+_j, c^-_j\rangle\}_{j=1}^M$  of triples, each contains a point $p_j$, the positive counter $c^+_j$ and a (sample) negative counter $c^-_j$. 

Random negative sampling strategy was used to train Bipolar-encoder. We used Adam optimizer to train, the batchsize of 4 and the learning\_rate of 3e-6. For Triplet loss part, the margin was set to be 1.0 and p was set to 2.  \\

\subsubsection{Unipolar-cls}
\begin{figure}
  \centering
  \includegraphics[width=0.45\textwidth]{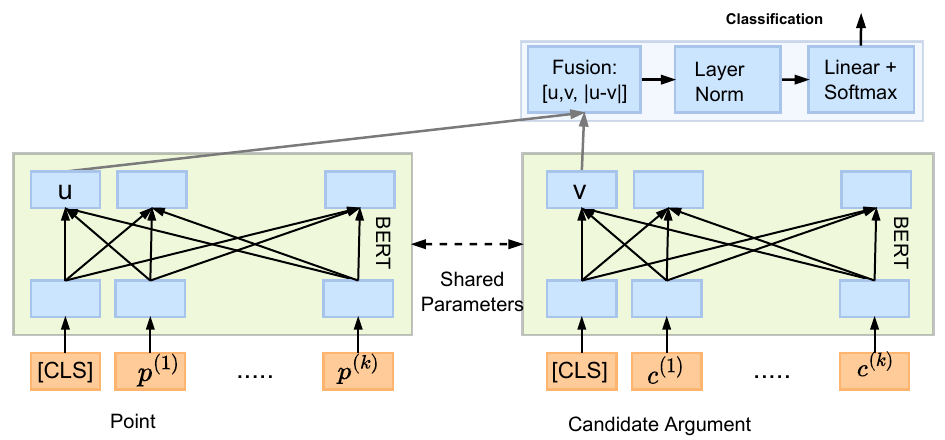}
 \caption{Unipolar-cls Architecture}
 \label{fig:unipolar-cls}
\end{figure}

Unipolar-cls is the simplified variant of Bipolar-encoder with only classification head. The architecture of Bipolar-cls is shown in Figure \ref{fig:unipolar-cls}. Bipolar-cls is trained based on cross-entropy loss on the dataset $\mathcal{D}^{point}=\{(\langle p_j,c_j\rangle,y_j)\}_{j=1}^M$ where $y_j$ is $1$ if $c_j$ is the correct counter-argument for the point $p_j$, and $0$ otherwise.

Random negative sampling strategy was used to train Bipolar-encoder. We used Adam optimizer to train, the batchsize of 4 and the learning\_rate of 3e-6. 

\subsubsection{Bipolar-encoder}

Random negative sampling strategy was used to train Bipolar-encoder. As stated in the main text, we trained the retrieval head and the classification head jointly. We used Adam optimizer to train, the batchsize of 4 and the learning\_rate of 3e-6. For Triplet loss part, the margin was set to be 1.0 and p was set to 2.  \\

\end{document}